\documentclass[review]{elsarticle}

\usepackage{float}
\usepackage{hyperref}
\usepackage{amsmath}
\usepackage{graphicx}
\usepackage{booktabs,dcolumn,caption}
\usepackage{blindtext, graphicx}
\usepackage[ruled,vlined,linesnumbered]{algorithm2e}
\usepackage{multirow}
\usepackage{amsmath,amssymb,amsfonts}
\usepackage[perpage]{footmisc}
\usepackage{caption}
\usepackage{subcaption}

\journal{arxiv}

\begin{document}

\begin{frontmatter}

\title{Hybrid Forest: A Concept Drift Aware Data Stream Mining Algorithm}

\author[AUTaddress]{Radin Hamidi Rad}
\ead{radin.h@aut.ac.ir}

\author[AUTaddress]{Maryam Amir Haeri}
\ead{haeri@aut.ac.ir}

\address[AUTaddress]{Department of Computer Engineering and Information Technology, Amirkabir University of Technology, Tehran, Iran}

\begin{abstract}
Nowadays with a growing number of online controlling systems in the organization and also a high demand of monitoring and stats facilities that uses data streams to log and control their subsystems, data stream mining becomes more and more vital. Hoeffding Trees (also called Very Fast Decision Trees a.k.a. VFDT) as a Big Data approach in dealing with the data stream for classification and regression problems showed good performance in handling facing challenges and making the possibility of any-time prediction. Although these methods outperform other methods e.g. Artificial Neural Networks (ANN) and Support Vector Regression (SVR), they suffer from high latency in adapting with new concepts when the statistical distribution of incoming data changes. In this article, we introduced a new algorithm that can detect and handle concept drift phenomenon properly. This algorithms also benefits from fast startup ability which helps systems to be able to predict faster than other algorithms at the beginning of data stream arrival. We also have shown that our approach will overperform other controversial approaches for classification and regression tasks.
\end{abstract}


\end{frontmatter}

\section{Introduction}

Hoeffding Trees (also called Very Fast Decision Trees a.k.a. VFDT) as a Big Data approach in dealing with the data stream for classification and regression problems showed good performance in handling facing challenges and making the possibility of any-time prediction. VFDTs are one of the most famous approaches in dealing with high-speed data streams with concept drift feature. VFDTs overcome other methods problems such as easy to be trapped in a local minimum (for ANN) and hard to decide proper kernel parameters and penalty (for SVR) \cite{Wu2016}, in contrast, they suffer from high latency in learning progress due to the existence of equally discriminative attributes. In this article, first, we address the cause of this latency and then introduce a method to diminish initial delay as much as possible.
One of the most realistic examples of data stream mining which illustrates the importance of this area is load forecasting or electricity price forecasting. Load forecasting is a crucial topic for power electricity suppliers and nowadays many data scientists are working on developing forecasting approaches to achieve good accuracy and robustness \cite{Ekici2016}, \cite{panamtash2018coherent} and \cite{Grolinger2016}. Another good example is fault detection in industries that help them to have a better guess of products life cycle so can achieve better maintenance services for their critical equipment \cite{Alzghoul2012}. Security is another field that needs data stream mining algorithms to maintain its reliabilities, sensor data and alarm notifications need to be analyzed to detect smart attack scenarios in moment \cite{amirhaerirteas}. In this article, we proposed a new data stream mining algorithm that not only covers all the weak points mentioned above but also improves the overall performance. We also uploaded our implementation for further use of other researchers. It can be reached here \footnote{\href{https://github.com/radinhamidi/Hybrid_Forest}{https://github.com/radinhamidi/Hybrid\_Forest}}. The main contribution of this paper are the following:
\begin{itemize}
  \item We proposed a novel algorithm for data stream mining proposed which is adaptable with concept drifts and can be used for both concept drift detection and data stream forecasting with the chance of having concept drifts.
  \item We discussed mathematical aspects of our proposed algorithm by details and proved that under certain conditions our approach works efficiently and guarantee to yields desirable output.
  \item We show that our new method has low computational complexity and can improve the prediction accuracy and over-perform some of the latest data stream mining systems.
\end{itemize}
In this article, we first discuss the related works in \autoref{sec:1} and address the issues each method faced to, then in \autoref{sec:2} we introduce our method which is an effort to resolve the discussed issued. Experimental results of implemented algorithms are available in \autoref{sec:3}. Finally, \autoref{sec:4} draws the final considerations and future work.

\section{Related Works} \label{sec:1}
CART or classification and regression tree \cite{Breiman1984} is among the most used machine learning methods because of its learning speed, accuracy, and overall performance \cite{Safavian1991}. Some key features like easy knowledge visualization, robustness and low chance of over-fit made this method preferable respect to other learning algorithms. There are lots of implementations of this algorithm for the different case of uses such as classification on batch data, classification on stream data, regression on batch data and finally regression on a stream of data. In this paper, we propose a new method for both regression and classification trees which operate on data stream.
Data mining on data stream faces challenges including concept drift and inability to access to all samples at any time. Aforementioned problems do not allow the learner to obtain overall vision on the whole set of data that cause a lack of certainty for choosing the best feature to split data by that. There are a couple of approaches that addressed the above issues. We will study them and other common methods briefly in the following paragraphs.
\newline\textbf{Very Fast Decision Tree}: VFDT has been announced by Domingos et al \cite{Domingos2000} in 2000. This algorithms has been specially designed for data streams and uses Hoeffding inequality to guarantee the right moment for turning a leaf to a splitting node. Consider a real-valued random variable $r$ whose range is $R$ (e.g., for a probability the range is one, and for an information gain the range is $R=\log c$, where $c$ is the number of classes). Suppose we have made $n$ independent observations of this variable and computed their mean $\bar{r}$. The Hoeffding bound states that, with the probability of $1-\delta$, the true mean of the variable is at least $\bar{r}-\varepsilon$, where:

\begin{equation} \label{hoeffding_eq1} 
\varepsilon = \sqrt{\frac{R^2\ln{(\frac{1}{\delta})}}{2n}}  
\end{equation}
The Hoeffding bound has a very attractive property that it is independent of the probability distribution generating the observations \cite{Domingos2000}. The price of this generality is that the bound is more conservative than distribution-dependent ones (i.e., it will take more observations to reach the same $\delta$ and $\varepsilon$) in return we will be able to determine the suitable amount of samples needed to start splitting the tree and making new decisions.
\newline\textbf{Hoeffding-Based Decision Trees with Options}: After introducing VFDT algorithm an intense concentration on data stream classification has been formed. On the original article by Domingos et al. \cite{Domingos2000}, an efficient approach for choosing appropriate feature has been introduced but after a while, another technical problem has been shown up which is fed from the inability of Hoeffding trees when two or more features have similar performance. In this situation, data samples will be accumulated on leaves but, respected leaves would not tend to split and leaf to child branches. In this particular problem, a wise solution introduced by Ikonomovska  et al. 2011 \cite{Ikonomovska2011}. According to the main article, an optional variable would be added to the Hoeffding tree which acts as an upper bound for the amount of accumulated data and chooses a feature for growing tree if samples over a specific leaf reach a threshold. However this solution may sound completely applicable at first sight but actually suffers from the problem of fine-tuning due to the fact that the thresholds vary from dataset to dataset. Also, this defined threshold may be dynamic and changes through data stream due to concept drift or similar phenomenon.
\newline\textbf{Random Forest of Very Fast Decision Trees}: Random forests can solve data variety and fit better on samples. This attitude let to implementing random forest of VDFTs which are a subclass of the decision trees \cite{Marron2014}. The fact that ensemble methods are always outperformed single algorithms is proved in Hensen \& Salamon 1990 \cite{Hansen1990}. They have uttered two condition for weak learners which guarantees this notion. A weak learner must be:
\begin{enumerate}
    \item \textit{Accurate: An accurate classifier is one that has an error rate of better than random guessing on new data values.}
    \item \textit{Diverse: Two classifiers are diverse if they make a different error on new data points}
\end{enumerate}
\textbf{Eraly Drift Detection Method}: EDDM which is introduced by Baena el al. \cite{baena2006early} is a stochastic based approach for drift detection. In this approach, system uses two thresholds to detect abrupt and gradual concept drifts. These thresholds leverage from famous three sigma rule that guarantees almost all of the distributed samples are not far from the mean point more than three times of standard deviation value. But this criterion lays on an important assumption: samples follow Gaussian distribution. As we know it is not mandatory for a data stream to follow a certain type of statistical distribution and also we may not aware or its distribution characteristics at all. Therefore, this approach may face poor performance for a range of datasets.
\newline\textbf{Support Vector Machine Approaches}: Using a SVM as a solution for data stream mining is discussed in various articles including \cite{liu2016svr} and \cite{he2011incremental}. Despite different proposed solutions to achieving an acceptable result, it can be seen that all of the approaches need kernel parameters to be set and other settings get done before starting. These pre-processing configurations make SVM/SVR approaches vulnerable to concept drifts.
\newline\textbf{Neural Network Approaches}: Neural Networks is one of the favorite solutions for sequential data mining. Recurrent neural networks and LSTMs are common approaches for sequential data mining. However, data streams are slightly different in their definition from sequential data, but these neural structures sometimes will be used as a solution \cite{leite2010evolving} \cite{DBLP:journals/corr/CuiSAH15}. The most important problem with these solutions is their famous weak point about the local minima. That behavior makes them not suitable for cases with concept drifts or situations that we know our problem is non-convex.
\newline In the rest of the paper, we discuss about our proposed approach and report the results of implementing random forest of VFDT and then introduce a brand new method for combining trees which not only improves the overall accuracy but also has significant effect on converge speed of single VFDT that can be an alternative solution beside Hoeffding trees with options that has its own difficulties and cons.

\section{Hybrid Forest of Very Fast Decision Trees} \label{sec:2}
\begin{figure}[ht]
\centerline{\includegraphics[width=\linewidth]{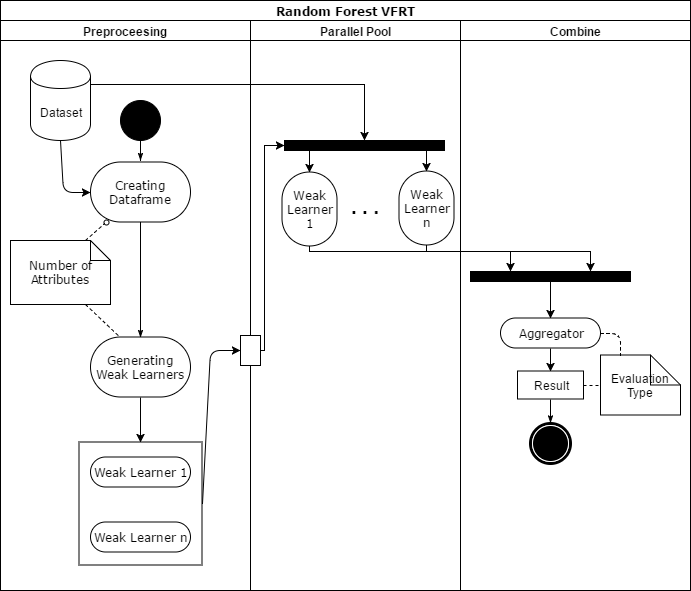}}
 \caption {Structure of the random forest algorithm containing bag of weak learners.}
\label{fig:1}
\end{figure}

As we discussed in related work section, an ensemble approach will always outperform single learning method under a certain condition, therefore, we started developing a random forest system consisting of Hoeffding trees and checked the results on famous datasets which are mentioned in dataset part in \autoref{sec:4} (Experimental Result). We used the VFML \cite{Hulten2003} engine as our main core for the regression task and a self-developed version for the classification task. The overall structure of the proposal system is shown in \autoref{fig:1}.
 
As can be seen in \autoref{fig:1}, the first version of random forest consist of 3 stages. First, the algorithm uses created data frame to obtain the number of features then generates weak learners. Each weak learner includes $\sqrt{(f_t)}$ features, where $f_t$ is length of our feature vector in data set. After generating weak learners, they will be sent to pool stage where algorithm ought to run all the given learners simultaneously and the result will be harvested after each data arrival and will be sent to combine stage where a judge rule will compute the final guess according to the evaluation type.
\subsection{Procedure}
After investigating random forest of VFDT we noticed a significant improvement in forecasting at the beginning of the data stream arrival. This phenomenon is the side effect of using weak learners as judgment for classification\textbackslash regression. In fact, by using the learner with less attribute than the main tree, the robustness of learners against variance of incoming data has decreased which cause learners to split data gathered at leaves quicker in compare to the complete tree, therefore, using random forest lead to having less error at the beginning. As a result, by asking those with less access to the whole feature set and using techniques like majority vote we can increase the speed of data mining at the arrival of a new data stream or occurrence of a new data distribution. This means we can improve mining speed at start-up and after each concept drift. This improvement unfortunately, has a side effect that needs to be handled and will be discussed in the rest.
Although adding more weak learners seems a solution for certain problems that have already mentioned, using ensemble method cannot yields to an acceptable result in compare with VFDT or other methods all the time. By arriving more samples, the complete tree slightly achieves better performance than ensemble version. That can be result of lack of adequate information for weak learners or constraint on depth of their trees due to feature vector size.

\begin{figure}[ht]
\centerline{\includegraphics[width=\linewidth]{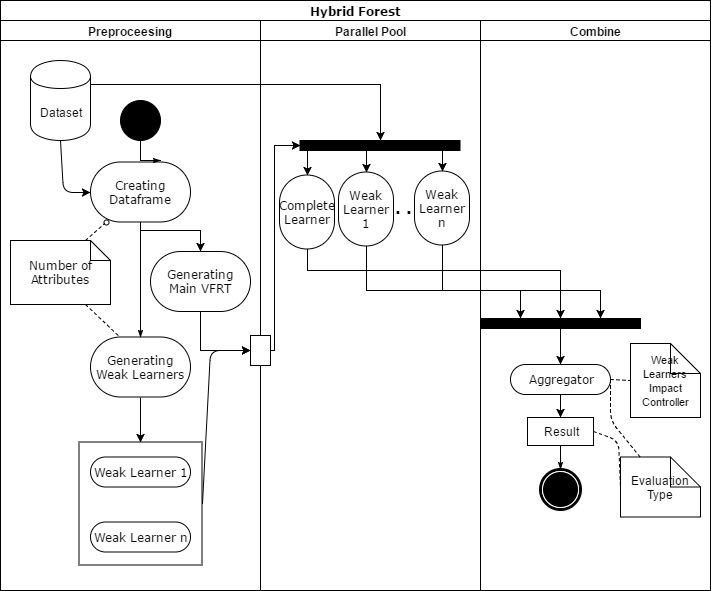}}
 \caption {Final version of proposed algorithm containing one full learner and bag of weak learners.}
\label{fig:2}
\end{figure}

Considering above observations about random forest decision trees we designed a new random forest under the name “Hybrid Forest” which includes a complete Hoeffding Tree as the main learner and a bunch of weak learners to improve forecasting in needed moments. The resulting structure of the system is shown in \autoref{fig:2}.

As it can be seen in \autoref{fig:2} there is a major difference between our proposed algorithm and random forest one that brought in \autoref{fig:1}. First, at the preprocessing stage, weak learners will be generated similar to the way they have been generated in the discussed method in figure 1. Second, the bag of weak learners and one full learner which has all the given attributes in itself will be sent to the pool step where algorithm ought to run all the given learners simultaneously and the result will be harvested after each data arrival and will be sent to combine stage where a judge rule will compute the final guess according to the weak learners impact controller unit and also evaluation type.
Due to the fact that the weak learners have not complete vision on the whole concept of dataset involving them in the mining of data stream in rest of the process effects on the accuracy of the system and reduce the overall performance. We solved this problem in "Weak Learner Impact Controller" part in combine stage. In this unit, we determine how weak learners should affect the final decision. The proposed unit is using a threshold
that will decide whether performs ensemble method or use the main tree solely. This approach is explained separately in \autoref{sec:2.3}.

\subsection{Efficient Number Of Weak Learners For Optimizing The Algorithm}\label{sec:2.2}
Replacing Single VFDT with an ensemble version of it can cause more use of both memory and computation resource of the machine which we name it computation cost here. In order to minimize the computation cost, it's needed to determine the efficient state for the proposed algorithm. We will show that under certain constraints for our random forest creation we can guarantee that our random forest members contain all the features at least once and therefore, our desired phenomenon of concept drift detection by weak learners will show up consequently. Here are defined constraints of our problem:
\begin{enumerate}
    \item For each attribute, there must be at least on weak learner among the group of weak learners that has it inside its bag of features.
    \item For convenience, we assumed that the number of features for each weak learner derive from thumb rule in \autoref{number_of_features}.
\end{enumerate}

\begin{equation} \label{number_of_features} 
 f_{wl} = \sqrt{f_{total}}
\end{equation}

Where $f_{wl}$ is the size of feature vectors for each weak learner and $f_{total}$ is the size of a complete feature vector of our system.

Therefore, if we shoe probability of not being chosen by none of the weak learners by $P$, then we can calculate $P$ as follows:
\begin{equation} \label{eq_p} 
 p = \binom{\sqrt{n}}{0} (\frac{1}{n})^0 (\frac{n-1}{n})^{\sqrt{n}}
\end{equation}
Where $n$ is size of feature vector for data stream. Above equation can be simplified and be written as:
\begin{equation} \label{eq_p_simplified} 
 p = (\frac{n-1}{n})^{\sqrt{n}}
\end{equation}
Now that we calculated the probability of not being chosen at all, we can calculate the probability of being chosen at least once by simply subtracting it from 1. So the probability of a feature to be chosen at least once will be:
\begin{equation} \label{eq_1_p} 
 q = 1 - p
\end{equation}
Where q is representation of probability of a feature to be chosen at least once. Since procedure of random forest creation consist of repetitive selection of features for each weak learner we must consider number of total selection times in our calculation. Let take $m$ as number of total weak learners and $n$ size of the original feature vector, then the probability of not being shown up ever during whole process of feature selection would be:
\begin{equation} \label{eq_p_m} 
\binom{\sqrt{m}}{0} (q)^0 (p)^m = p^m
\end{equation}

\begin{equation} \label{eq_p_final} 
p_{\text{ not observed}} = ({\frac{n-1}{n}})^{m\sqrt{n}}
\end{equation}

Therefore, as the \autoref{eq_p_final} utters, the probability of not being chosen even once in $m$ times random with replacement selection of features for weak learners is equal to $({\frac{n-1}{n}})^{m\sqrt{n}}$. Now we can select the desired size of our random forest according to $n$ which is the size of the original feature vector and calculated probability of presentation of each feature at least once. For better intuition of the current equation, it's better to subtract the probability from one to obtain the probability of being seen at least once since it's our final desired value. The final form of the desired variable for our work is shown in \autoref{desired_prob}.

\begin{equation} \label{desired_prob} 
P_{confidence} = 1 - ({\frac{n-1}{n}})^{m\sqrt{n}}
\end{equation}

\begin{figure}[H]
\begin{subfigure}{1\textwidth}
\centering
    \includegraphics[width=1\textwidth]{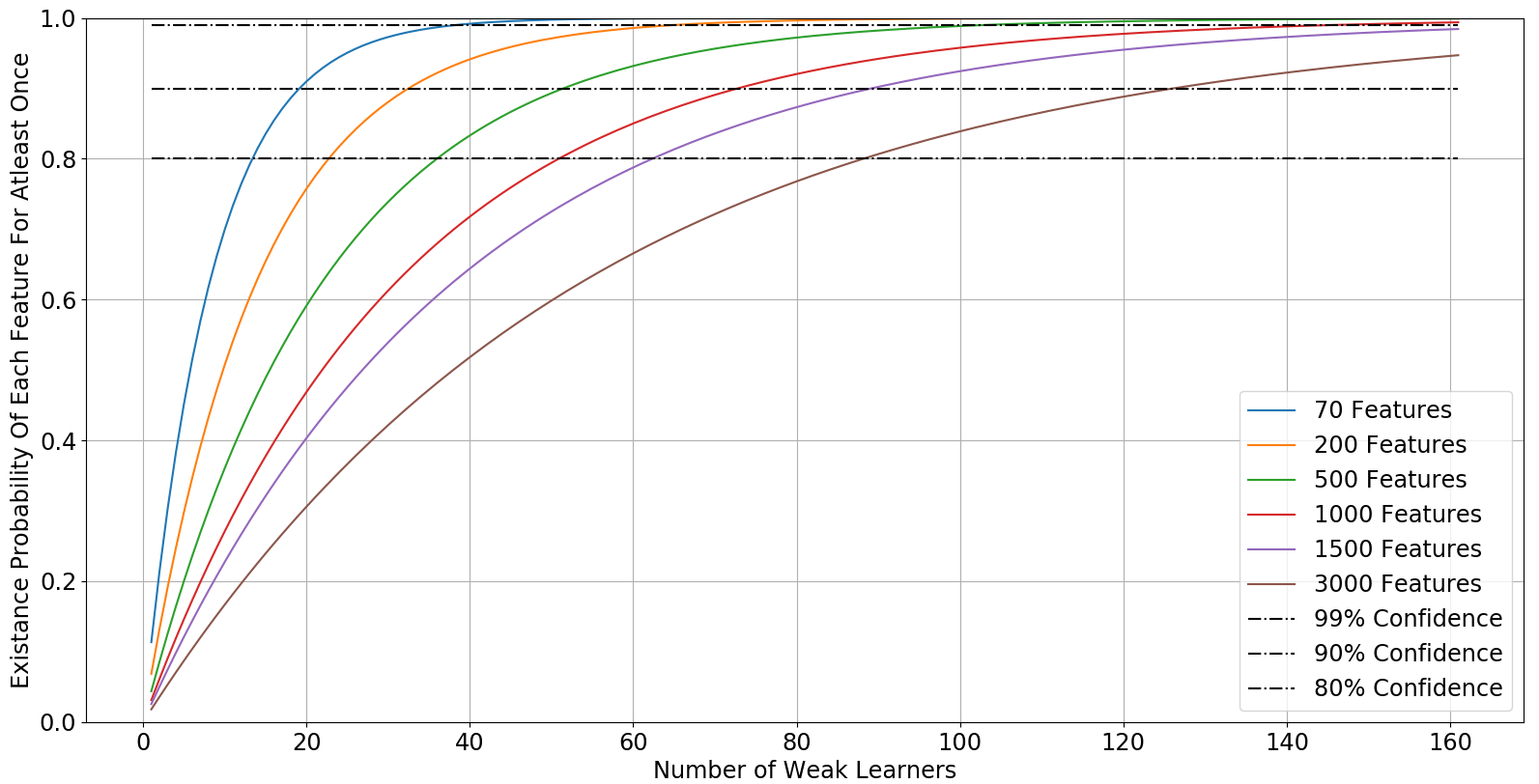}
 \caption {Confidence for different number of features and weak learners.}
\label{fig:3a}
\end{subfigure}
\begin{subfigure}{1\textwidth}
\centering
    \includegraphics[width=1\textwidth]{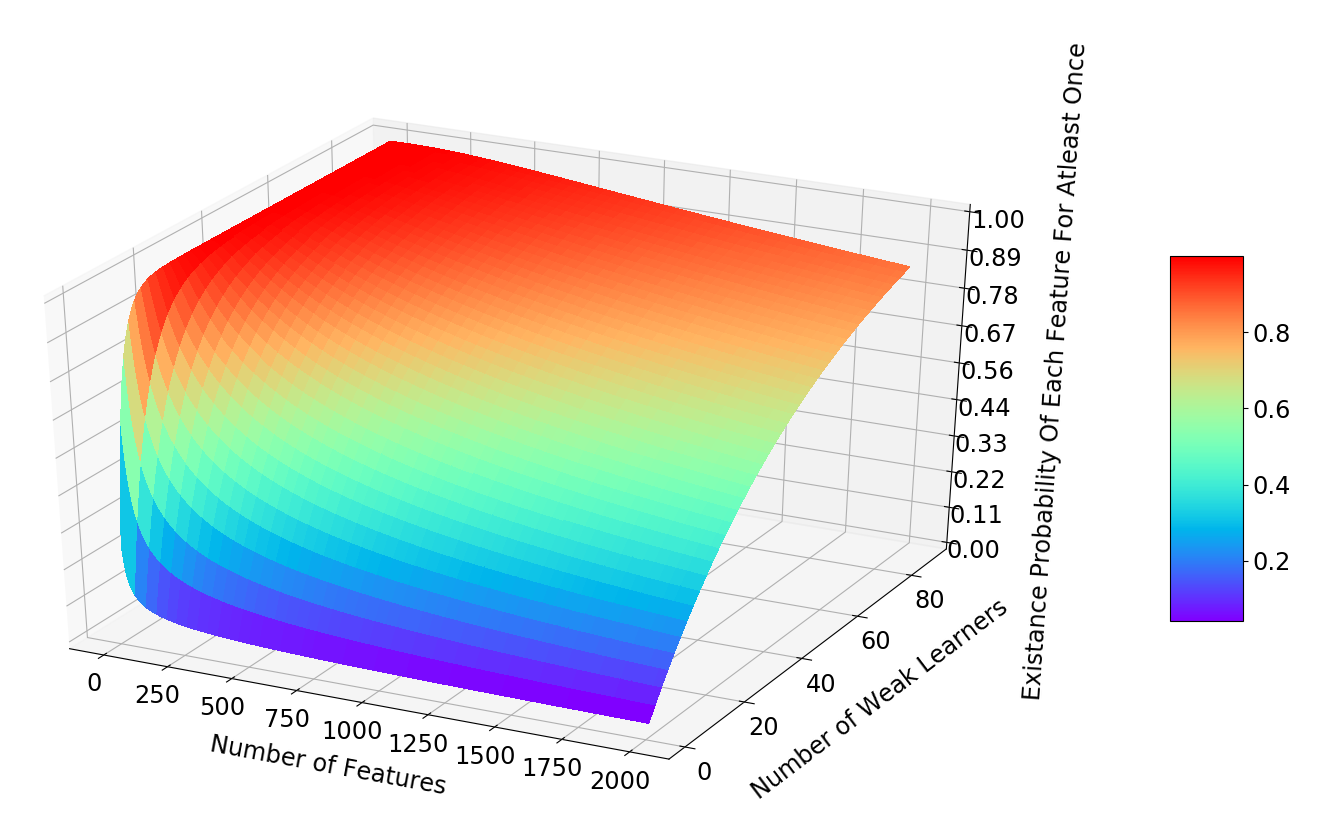}
 \caption {3D representation of confidence for different number of features and weak learners.}
\label{fig:3b}
\end{subfigure}
\caption{We can see how the number of weak learners and size of feature vector can influence the probability.}
\label{fig:3}
\end{figure}

For better inference, our proposed confidence equation is calculated for a set of the feature vector with a different number of weak learners and rendered in \autoref{fig:3}. This curve can be handy for learners community size selection and saves the time.

It is clear that by increasing $m$ which is the number of weak learners, the confidence will approach to its maximum value that is one and by decreasing the size of the feature vector our confidence will grow faster that means needs less number of weak learners so dimensionality reduction and removing redundant features may cause better performance significantly. Furthermore, we can clearly see that confidence grow nonlinear, thus even systems with low computational capabilities can benefit from the algorithm by adding a few weak learners.

\subsection{Weak Learners Impact Controller}\label{sec:2.3}
As we discussed above, we need a system to decide how weak learner affect the decision-making process. This algorithm is proposed in "Weak Learner Impact Controller" which is shown in \autoref{fig:2}. The decision making algorithm uses a window to accumulate the last $k_{\text{window size}}$ results from each weak learner and if more than $d_{\text{hybrid impact}}$ of the window has accumulated with correct results then we use the majority vote for decision making else we just use the main tree. The mentioned algorithm is brought in \autoref{algorithm:1}. Please notice that variables that are written in uppercase refer to arrays. $X_{wl}$ is the array of outputs for weak learners, $x_{t}$ is the output of the main Hoeffding tree, $W$ is the window that stores performance of past decisions. The window array ($W$) is a FIFO buffer that stores $k_{\text{window size}}$ last performance and with every new sample arrival, the oldest record will be deleted and a new record will be added to the end of the buffer.

\bigskip 
\begin{algorithm}[H]
\DontPrintSemicolon
\SetAlgoLined
\KwResult{Determined Class or Forecasted Value}
\SetKwInOut{Input}{Input}\SetKwInOut{Output}{Output}
\Input{$X_{wl}$ , $x_{t}$, $W$, $d_{\text{hybrid impact}}$, $k_{\text{window size}}$, $Y_{Actual}$}
\Output{$Y_{Predicted}$}
window is filled with ones initially \;
\BlankLine
\ForEach{ incoming sample}{    
    \BlankLine
    \tcc{Deciding}
    \uIf{Sum($W$) \textgreater ($d_{\text{hybrid impact}}\times k_{\text{window size}}$)}{
        $Y_{Predicted}$ = MajorVote($X_{wl}$ , $x_{t}$)}
    \uElse{
        $Y_{Predicted}$ = $x_{t}$
    }
    \tcc{Updating}    
    \uIf{$Y_{Predicted}$ == $Y_{Actual}$}{
        $W\leftarrow1$}
    \uElse{
        $W\leftarrow0$
    }
}
 \caption{Hybrid impact controller}
 \label{algorithm:1}
\end{algorithm}

\subsection{Concept Drift Detection}\label{sec:2.4}
Now that we have a log of past performance of weak learners and complete Hoeffding tree, we can do concept drift detection task. Concept drift detection can be done by comparing the performance of single Hoeffding tree with weak learners through the window and if for a repeated number of times weak learners work correctly while single complete tree returns the wrong answer we can realize that a concept drift has happened. This approach has benefits in comparison to other controversial methods. First, this approach does not follow any preset criterion like a specific probability distribution. Also, this method is sensitive to both gradual and abrupt concept drift and can be used for both purposes. Finally, this method does not need any further variable to be stored and previous variables that are used in the algorithm are enough for this task to be done. 

\section{Experimental Results} \label{sec:3}

In this section, we examine our method based on famous and trusted datasets that are used in related works and try to clarify how much the impact will be if our proposed algorithm is used.
\subsection{Data Set}
Due to the fact that best illustration will be achieved when we check any-time accuracy and converge speed on the same datasets that are used on previous methods, we have tested our algorithm on two “benchmark” datasets (wind and abalone) from different famous dataset repositories, the UCI Machine Learning Repository (Frank \& Asuncion, 2010) \cite{Dua:2017} and the Delve Repository. For the classification task evaluation, we selected a dataset with high feature vector dimension for better intuition. As it's discussed in \autoref{sec:2.2} number of weak learners is directly related to the feature vector dimension. Waveform dataset from UCI Machine Learning Repository\cite{Dua:2017} with 21 features and 3 classes is selected for the classification task.

\subsection{Classification Results and Discussion}
In this section, we will discuss the results of the proposed algorithm and will analyze its performance. Hybrid forest implementation can be found on its Github \footnote{\href{https://github.com/radinhamidi/Hybrid_Forest}{https://github.com/radinhamidi/Hybrid\_Forest}} page. The result of repeated runs on waveform dataset for three different methods including the proposed model can be found in \autoref{tab:1}. These results are reported after running algorithms for 20 times repeatedly.

\begin{table}[H]
\centering
\captionsetup{justification=centering,
              labelsep=newline,
              singlelinecheck=false,
              skip=0.333\baselineskip}
\caption{Accuracy of different methods for $d_{\text{hybrid impact}} = 0.2$ and $k_{\text{window size}} = 15$}\label{tab:1}
\begin{tabular}{|c|l|l|l|}
\hline
\multicolumn{1}{|l|}{Data Set} & Method        & Accuracy & Standard Deviation \\ \hline
\multirow{3}{*}{Waveform}      & VFDT          & 72.5\%   & 0.0                \\ \cline{2-4} 
                               & Random Forest & 76.3\%   & 1.57               \\ \cline{2-4} 
                               & Hybrid Forest & 77.73\%  & 1.19               \\ \hline
\end{tabular}%
\end{table}

\begin{figure}[H]
\begin{subfigure}{.5\textwidth}
\centering
\includegraphics[width=\linewidth]{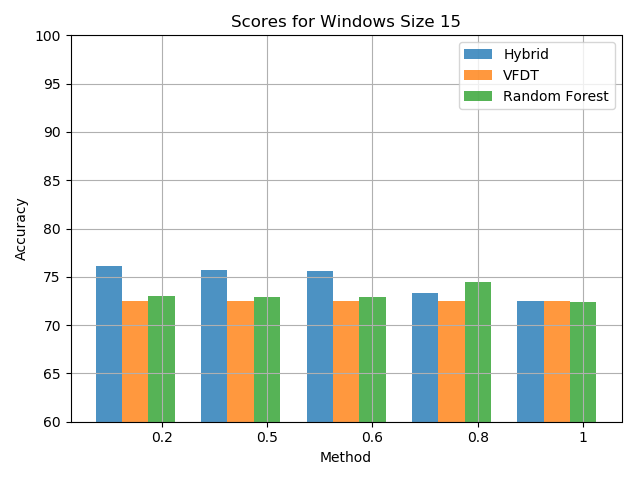}
\caption {Fix window size = 15}
\label{fig:7a}
\end{subfigure}
\begin{subfigure}{.5\textwidth}
\centering
\includegraphics[width=\linewidth]{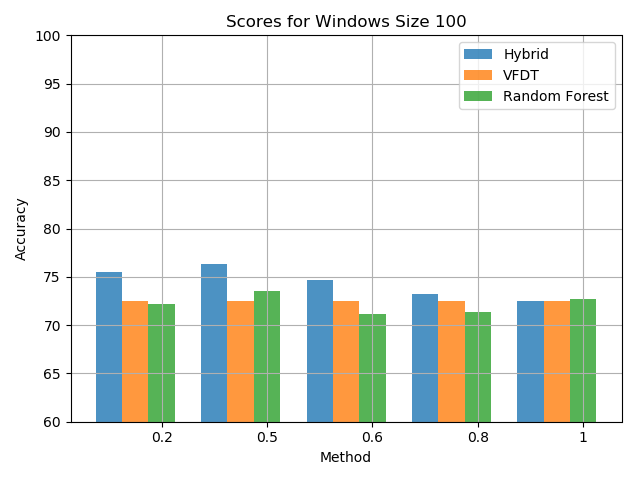}
 \caption {Fix window size = 100}
\label{fig:7b}
\end{subfigure}
\begin{subfigure}{.5\textwidth}
\centering
\includegraphics[width=\linewidth]{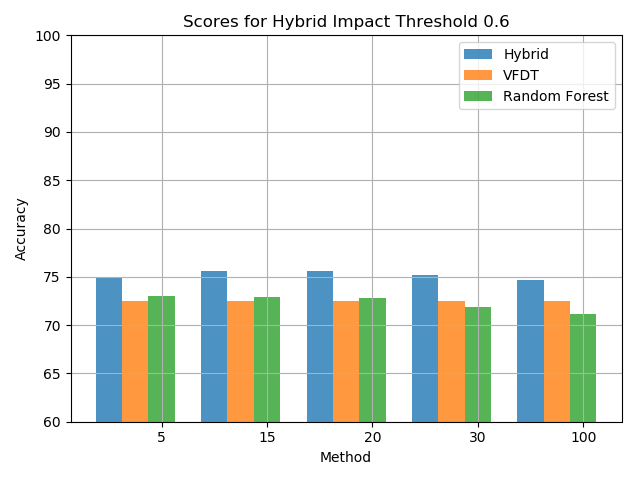}
 \caption {Fix hybrid impact = 0.6}
\label{fig:7c}
\end{subfigure}
\begin{subfigure}{.5\textwidth}
\centering
\includegraphics[width=\linewidth]{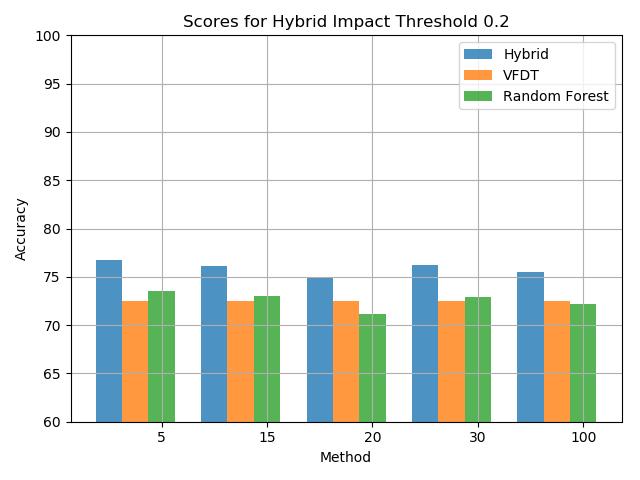}
 \caption {Fix hybrid impact = 0.2}
\label{fig:7d}
\end{subfigure}
\caption{Comparison of Accuracy between single Hoeffding tree, Hybrid forest and random forest for different configurations.}
\label{fig:7}
\end{figure}

As it can be seen in \autoref{fig:7}, hybrid forest over-perform both random forest and VFDT. The result is brought from a different configuration for better inference. Rest of the plots can be found in its repository page.

\subsection{Windows Size and Hybrid Impact Analysis}
\begin{figure}[H]
\begin{subfigure}{.5\textwidth}
\centering
\includegraphics[width=\linewidth]{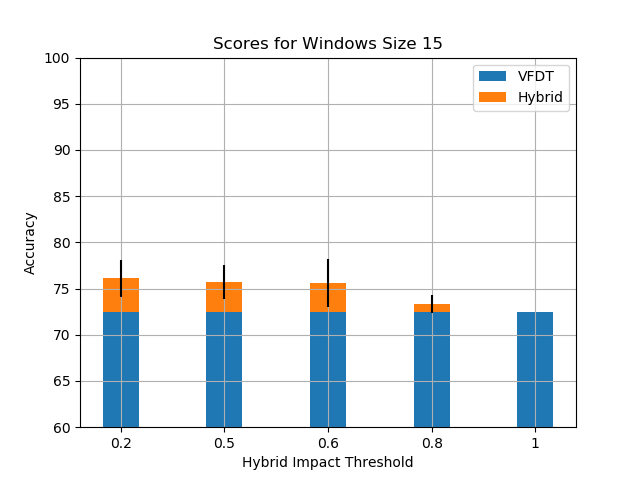}
\caption {Fix window size = 15}
\label{fig:8a}
\end{subfigure}
\begin{subfigure}{.5\textwidth}
\centering
\includegraphics[width=\linewidth]{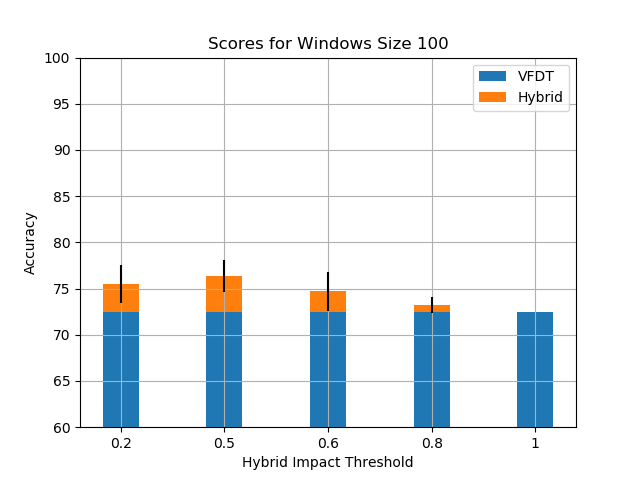}
\caption {Fix window size = 100}
\label{fig:8b}
\end{subfigure}
\begin{subfigure}{.5\textwidth}
\centering
\includegraphics[width=\linewidth]{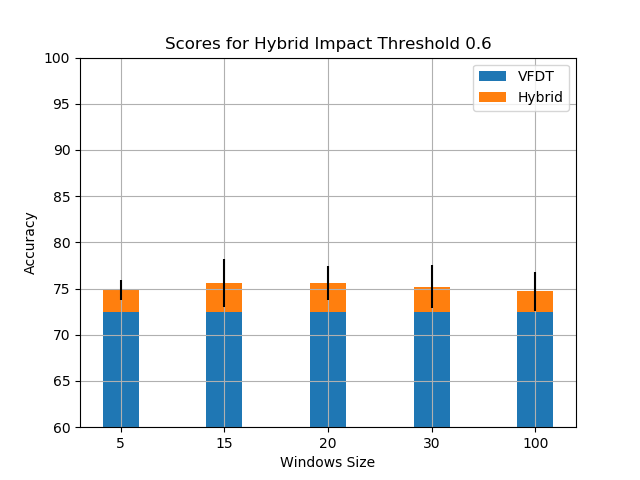}
 \caption {Fix hybrid impact = 0.6}
\label{fig:8c}
\end{subfigure}
\begin{subfigure}{.5\textwidth}
\centering
\includegraphics[width=\linewidth]{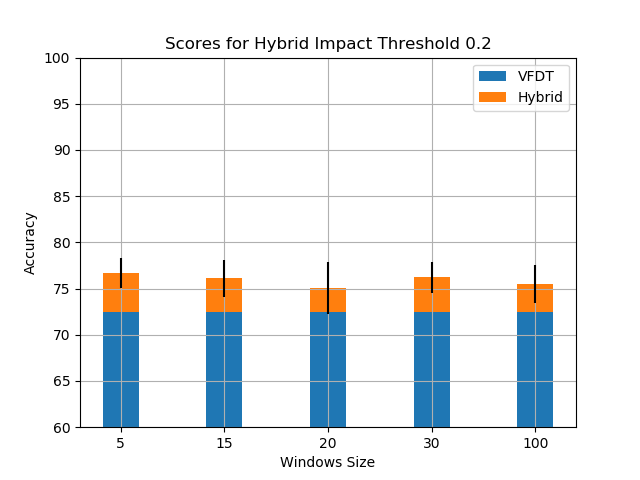}
 \caption {Fix hybrid impact = 0.2}
\label{fig:8d}
\end{subfigure}
\caption{Comparison of accuracy between single Hoeffding tree and Hybrid forest for different configurations.}
\label{fig:8}
\end{figure}

However fine-tuning can help algorithms to have better performance for each situation, they are annoying and sometimes will be considered as a negative point for an approach. In hybrid forest approach, $d_{\text{hybrid impact}}$ and $k_{\text{window size}}$ are two parameters that should be set before the start. In this section, we will discuss the usage of these two variables and suggest an estimation thumb rule for those users who want to use the hybrid forest. Starting with $k_{\text{window size}}$, this parameter represents that memory that algorithm uses to determine whether if it is useful to weak learners result or not. Each data stream has its own frequency and sample input rate and this rate is part of its instinct characteristics. This fact is the main cause of using $k_{\text{window size}}$, to get more adapted with a different data stream. As a result, we could find out our proper window size just by checking the problem's characteristics. The other parameter, $d_{\text{hybrid impact}}$ is the one and only control that represents the trade-off between pure random forest approach and single complete tree approach. Typically a user can determine proper value by checking the feature vector richness and efficiency. Problems with more features should rely more on the ensemble aspect of the hybrid forest, therefore, $d_{\text{hybrid impact}}$ should be close to zero. In contrast, if the problem suffers from lack of features $d_{\text{hybrid impact}}$ should be close to the one to achieve desirable results.
As it is shown in \autoref{fig:8}, different values for the proposed algorithm can yield different performance. For a better understanding of how these parameters affect the result, we brought some results in \autoref{fig:8}. Please note that all the experiments here are the mean value of 20 separate tests. Rest of the plots can be found in its repository page.

\subsection{Concept Drift}
\begin{figure}[H]
\centerline{\includegraphics[width=1\linewidth]{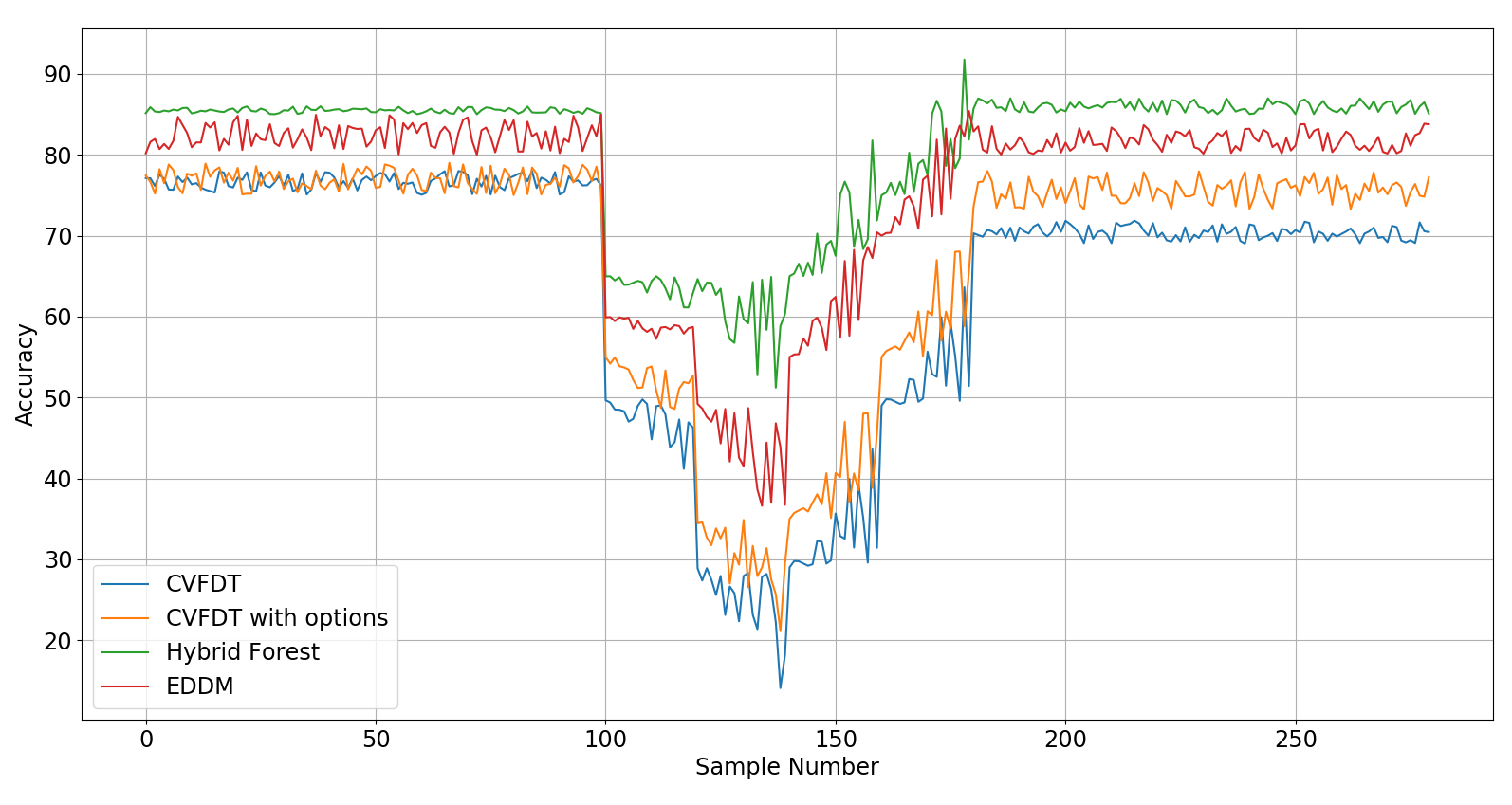}}
 \caption {Concept drift detection and recovery comparison}
\label{fig:concept_drift}
\end{figure}

This section represents experimental results of concept drift detection solution that were introduced in \autoref{sec:2.4}. For a better comparison over all methods, we compared the result of concept drift detection and recovery after that on four different approaches: CVFDT, CVFDT with options, EDDM-VFDT, and Hybrid Forest. As it is shown in \autoref{fig:concept_drift}, we can see that the proposed algorithm can better handle the concept drift and lose less accuracy. That is because, when the main tree falls in wrong track of stream and cannot follow that correctly, the weak learners will start alarming and switch output values from a single tree to the ensemble mode.

\subsection{Regression Results And Discussion}
While the Hybrid Forest algorithm starts to converge to real values very fast, the original Hoeffding tree cannot follow the changes correctly and causes a severe error at start-up. To demonstrate how effective using hybrid method could be we brought the results of forecasting in mean accuracy format for both datasets in \autoref{tab:2}. The results show how many samples are needed by mentioned algorithms to reach the $90\%$ accuracy and keep track of the input data stream.


\begin{table}[h]
\centering
\captionsetup{justification=centering,
              labelsep=newline,
              singlelinecheck=false,
              skip=0.333\baselineskip}
\caption{Speed Of convergence at start-up for different algorithms}\label{tab:2}
\begin{tabular}{|c|c|c|}
\hline
\multicolumn{3}{|c|}{Number of Samples}              \\ \hline
\multirow{3}{*}{Abalone} & Single VFRT        & 1042 \\ \cline{2-3} 
                         & Random Forest VFRT & 185  \\ \cline{2-3} 
                         & Hybrid Forest VFRT & 185  \\ \hline
\multirow{3}{*}{Wind}    & Single VFRT        & 1073 \\ \cline{2-3} 
                         & Random Forest VFRT & 200  \\ \cline{2-3} 
                         & Hybrid Forest VFRT & 712  \\ \hline
\end{tabular}
\end{table}

As you can see Hybrid forest reaches the accuracy of $90\%$ faster than single VFRT. Both the Random Forest and Hybrid Forest act better than Single Tree. However, the Random Forest shows good performance in the beginning but it suffers from the high standard deviation in accuracy in long-term scope.
Following figures show major improvement in forecasting at the beginning of the flow of the data stream.
 
\begin{figure}[H]
\centerline{\includegraphics[width=\linewidth]{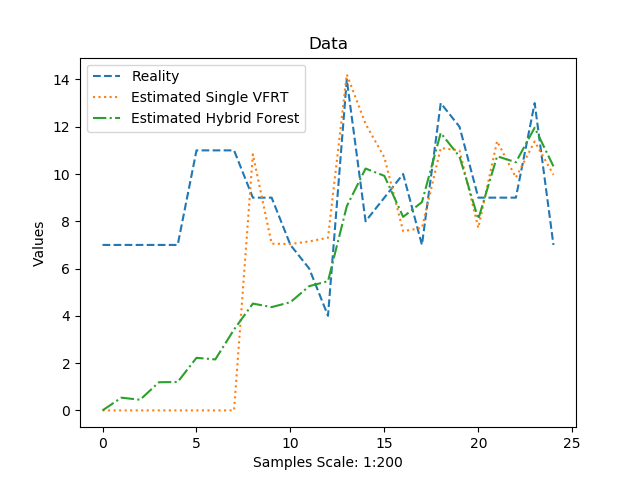}}
 \caption {Converging comparison on \textbf{abalone} data set.}
\label{fig:4}
\end{figure}
 
As is shown in \autoref{fig:4}, Hybrid tree method starts converging faster than single Hoeffding regression tree. This behavior of the hybrid tree is because of an intrinsic feature of weak learners that grow faster than a complete tree and make it possible to estimate at the very beginning of data flow.

\begin{figure}[H]
\centerline{\includegraphics[width=\linewidth]{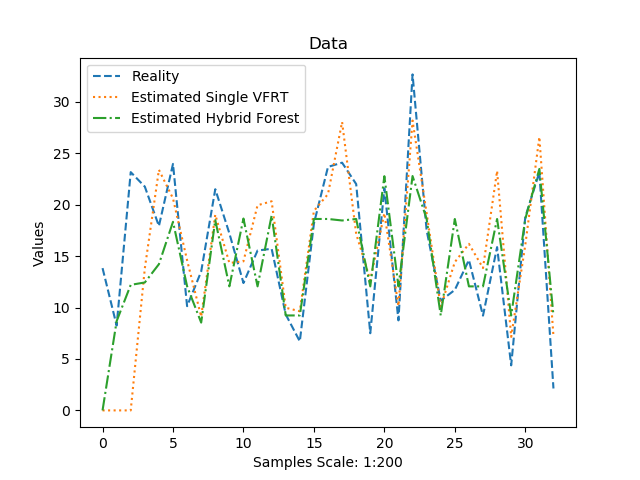}}
 \caption {Converging comparison on \textbf{wind} data set.}
\label{fig:5}
\end{figure}
 
Also \autoref{fig:5} shows the same behavior of the proposed system on the different dataset. It’s good to know that for convenience in compare we downsampled the estimation results with 1:200 ratio.
 
\begin{figure}[H]
\begin{subfigure}{.5\textwidth}
\centering
\includegraphics[width=\linewidth]{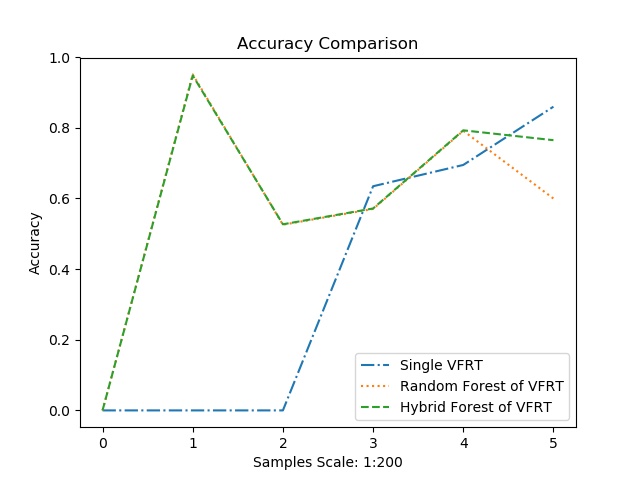}
 \caption {\textit{abalone} dataset}
\label{fig:6a}
\end{subfigure}
\begin{subfigure}{.5\textwidth}
\centering
\includegraphics[width=\linewidth]{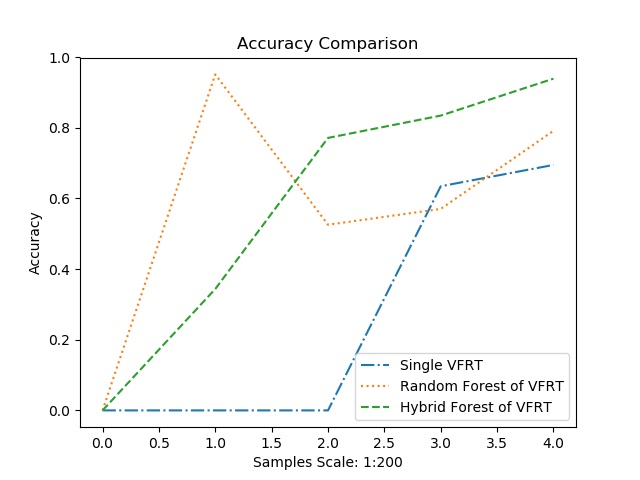}
 \caption {\textit{wind} dataset}
\label{fig:6b}
\end{subfigure}
\caption{Comparison of accuracy between single Hoeffding regression tree, random forest of it and our proposed method called “Hybrid Forest”.}
\label{fig:6}
\end{figure}
 
 
As shown above, \autoref{fig:6} compares the accuracy of our method versus a single VFRT and also random forest of VFRT. As it can be seen both hybrid and random forest methods which are ensemble approaches outperform the single tree one. In contrast, hybrid forest uses less memory than a random forest since needs less weak learners. To wrap these results up we can conclude that with consuming a few more memory than single tree we can achieve a fast start-up for our regression task without losing concept by using a complete learner in our ensemble method.

\subsection{Memory Analysis}
In \autoref{tab:3} you can find the memory usage of each method separately.

\begin{table}[h]
\centering
\captionsetup{justification=centering,
              labelsep=newline,
              singlelinecheck=false,
              skip=0.333\baselineskip}
\caption{Memory usage table.}\label{tab:3}
\begin{tabular}{|c|c|c|}
\hline
\multicolumn{3}{|c|}{Memory Usage in (MB)}            \\ \hline
\multirow{3}{*}{Abalone} & Single VFRT        & 0.36  \\ \cline{2-3} 
                         & Random Forest VFRT & 9     \\ \cline{2-3} 
                         & Hybrid Forest VFRT & 9.36  \\ \hline
\multirow{3}{*}{Wind}    & Single VFRT        & 1.23  \\ \cline{2-3} 
                         & Random Forest VFRT & 30.75 \\ \cline{2-3} 
                         & Hybrid Forest VFRT & 31.98 \\ \hline
\end{tabular}
\end{table}

As the \autoref{tab:3} shows the difference of memory needed for implementing the method is reasonable and also reachable in most of today practical machines.
\textbf{Note:} Random forest and hybrid model are both contain 100 weak learners at the time logs are captured for \autoref{tab:3}. Thus, memory usage may vary from case to case.

\section{Conclusions} \label{sec:4}
This paper proposes a new method for classification and regression tasks on data streams based on Hoeffding Trees. Slow convergence at the beginning of data flow and concept drift is the most common issues with these trees. We proposed an ensemble algorithm and that can help to fix the addressed issues. This will also improve the accuracy in general. We also investigated the characteristics of our algorithm theoretically and discussed the behavior of each variable solely. Beside mentioned advantages, the introduced method does not take eye-catching computational resource in comparison to previous methods and they can shift to the proposed algorithm easily. For future work, other hybrid controller approaches can be researched to achieve a more accurate algorithm.


\bibliography{main}

\begin{thebibliography}{10}
\expandafter\ifx\csname url\endcsname\relax
  \def\url#1{\texttt{#1}}\fi
\expandafter\ifx\csname urlprefix\endcsname\relax\def\urlprefix{URL }\fi
\expandafter\ifx\csname href\endcsname\relax
  \def\href#1#2{#2} \def\path#1{#1}\fi

\bibitem{Wu2016}
X.~Wu, J.~He, T.~Yip, J.~Lu, N.~Lu, {A two-stage random forest method for
  short-term load forecasting}, IEEE Power and Energy Society General Meeting
  2016-Novem~(2015).
\newblock \href {http://dx.doi.org/10.1109/PESGM.2016.7741295}
  {\path{doi:10.1109/PESGM.2016.7741295}}.

\bibitem{Ekici2016}
S.~Ekici, {Electric Load Forecasting Using Regularized Extreme Learning
  Machines}, International Journal of Industrial Electronics and Electrical
  Engineering~(6) (2016) 119--122.

\bibitem{panamtash2018coherent}
H.~Panamtash, Q.~Zhou, Coherent probabilistic solar power forecasting, in: 2018
  IEEE International Conference on Probabilistic Methods Applied to Power
  Systems (PMAPS), IEEE, 2018, pp. 1--6.

\bibitem{Grolinger2016}
K.~Grolinger, A.~L'Heureux, M.~A. Capretz, L.~Seewald, {Energy forecasting for
  event venues: Big data and prediction accuracy}, Energy and Buildings 112
  (2016) 222--233.
\newblock \href {http://dx.doi.org/10.1016/j.enbuild.2015.12.010}
  {\path{doi:10.1016/j.enbuild.2015.12.010}}.

\bibitem{Alzghoul2012}
A.~Alzghoul, M.~L{\"{o}}fstrand, B.~Backe,
  \href{http://dx.doi.org/10.1016/j.cie.2011.12.023}{{Data stream forecasting
  for system fault prediction}}, Computers and Industrial Engineering 62~(4)
  (2012) 972--978.
\newblock \href {http://dx.doi.org/10.1016/j.cie.2011.12.023}
  {\path{doi:10.1016/j.cie.2011.12.023}}.
\newline\urlprefix\url{http://dx.doi.org/10.1016/j.cie.2011.12.023}

\bibitem{amirhaerirteas}
M.~AmirHaeri, R.~Jalili, Rteas: A real-time algorithm for extracting attack
  scenarios from intrusion alert stream.

\bibitem{Breiman1984}
L.~Breiman, J.~Friedman, R.~Olshen, C.~Stone, {Classification and Regression
  Trees}, Belmont, Calif.: Wadsworth.

\bibitem{Safavian1991}
S.~R. Safavian, D.~Landgrebe, {A Survey of Decision Tree Classifier
  Methodology}, IEEE Transactions on Systems, Man and Cybernetics 21~(3) (1991)
  660--674.
\newblock \href {http://dx.doi.org/10.1109/21.97458}
  {\path{doi:10.1109/21.97458}}.

\bibitem{Domingos2000}
P.~Domingos, G.~Hulten,
  \href{http://portal.acm.org/citation.cfm?doid=347090.347107}{{Mining
  high-speed data streams}}, Kdd (2000) 71--80\href
  {http://dx.doi.org/10.1145/347090.347107} {\path{doi:10.1145/347090.347107}}.
\newline\urlprefix\url{http://portal.acm.org/citation.cfm?doid=347090.347107}

\bibitem{Ikonomovska2011}
E.~Ikonomovska, J.~Gama, B.~Zenko, S.~Dzeroski,
  \href{http://machinelearning.wustl.edu/mlpapers/paper{\_}files/ICML2011Ikonomovska{\_}349.pdf}{{Speeding-up
  Hoeffding-based regression trees with options}}, Proceedings of the {\ldots}
  (2011) 537--544.
\newline\urlprefix\url{http://machinelearning.wustl.edu/mlpapers/paper{\_}files/ICML2011Ikonomovska{\_}349.pdf}

\bibitem{Marron2014}
D.~Marron, A.~Bifet, G.~{De Francisci Morales}, {Random forests of very fast
  decision trees on GPU for mining evolving big data streams}, Frontiers in
  Artificial Intelligence and Applications 263 (2014) 615--620.
\newblock \href {http://dx.doi.org/10.3233/978-1-61499-419-0-615}
  {\path{doi:10.3233/978-1-61499-419-0-615}}.

\bibitem{Hansen1990}
L.~K. Hansen, P.~Salamon, {Neural Network Ensembles}, IEEE Transactions on
  Pattern Analysis and Machine Intelligence 12~(10) (1990) 993--1001.
\newblock \href {http://dx.doi.org/10.1109/34.58871}
  {\path{doi:10.1109/34.58871}}.

\bibitem{baena2006early}
M.~Baena-Garc{\'\i}a, J.~del Campo-{\'A}vila, R.~Fidalgo, A.~Bifet,
  R.~Gavald{\`a}, R.~Morales-Bueno, Early drift detection method.

\bibitem{liu2016svr}
J.~Liu, E.~Zio, A svr-based ensemble approach for drifting data streams with
  recurring patterns, Applied Soft Computing 47 (2016) 553--564.

\bibitem{he2011incremental}
H.~He, S.~Chen, K.~Li, X.~Xu, Incremental learning from stream data, IEEE
  Transactions on Neural Networks 22~(12) (2011) 1901--1914.

\bibitem{leite2010evolving}
D.~Leite, P.~Costa, F.~Gomide, Evolving granular neural network for
  semi-supervised data stream classification, in: Neural Networks (IJCNN), The
  2010 International Joint Conference on, IEEE, 2010, pp. 1--8.

\bibitem{DBLP:journals/corr/CuiSAH15}
Y.~Cui, C.~Surpur, S.~Ahmad, J.~Hawkins,
  \href{http://arxiv.org/abs/1512.05463}{Continuous online sequence learning
  with an unsupervised neural network model}, CoRR abs/1512.05463.
\newblock \href {http://arxiv.org/abs/1512.05463} {\path{arXiv:1512.05463}}.
\newline\urlprefix\url{http://arxiv.org/abs/1512.05463}

\bibitem{Hulten2003}
G.~Hulten, P.~Domingos, \href{http://www.cs.washington.edu/dm/vfml/}{{VFML -- A
  toolkit for mining high-speed time-changing data streams}} (2003).
\newline\urlprefix\url{http://www.cs.washington.edu/dm/vfml/}

\bibitem{Dua:2017}
D.~Dheeru, E.~Karra~Taniskidou, \href{http://archive.ics.uci.edu/ml}{{UCI}
  machine learning repository} (2017).
\newline\urlprefix\url{http://archive.ics.uci.edu/ml}

\end{thebibliography}

\end{document}